\documentclass[11pt]{article}
\usepackage[final]{acl}

\usepackage{times}
\usepackage{latexsym}
\usepackage{tabularx}
\usepackage{tikz}
\usepackage{inconsolata}
\usepackage{booktabs}
\usepackage{multirow}
\usepackage{xcolor}
\usepackage{svg}
\usepackage{longtable}

\usepackage[T1]{fontenc}
\usepackage[utf8]{inputenc}

\usepackage{microtype}

\usepackage{inconsolata}

\usepackage{graphicx}

\title{Generalists vs. Specialists: Evaluating Large Language Models for Urdu}

\author{Samee Arif \and Abdul Hameed Azeemi \and Agha Ali Raza \\
        \{samee.arif, abdul.azeemi, agha.ali.raza\}@lums.edu.pk \\
        Lahore University of Management Sciences \AND
        Awais Athar \\ awais@ebi.ac.uk \\ EMBL European Bioinformatics Institute}

\begin{document}
\maketitle
\begin{abstract}
In this paper, we compare general-purpose models, \texttt{GPT-4-Turbo} and \texttt{Llama-3-8b}, with special-purpose models—\texttt{XLM-Roberta-large}, \texttt{mT5-large}, and \texttt{Llama-3-8b}—that have been fine-tuned on specific tasks. We focus on seven classification and seven generation tasks to evaluate the performance of these models on Urdu language. Urdu has 70 million native speakers, yet it remains underrepresented in Natural Language Processing (NLP). Despite the frequent advancements in Large Language Models (LLMs), their performance in low-resource languages, including Urdu, still needs to be explored. We also conduct a human evaluation for the generation tasks and compare the results with the evaluations performed by \texttt{GPT-4-Turbo}, \texttt{Llama-3-8b} and \texttt{Claude 3.5 Sonnet}. We find that special-purpose models consistently outperform general-purpose models across various tasks. We also find that the evaluation done by \texttt{GPT-4-Turbo} for generation tasks aligns more closely with human evaluation compared to the evaluation the evaluation done by $\texttt{Llama-3-8b}$. This paper contributes to the NLP community by providing insights into the effectiveness of general and specific-purpose LLMs for low-resource languages.

\end{abstract}

\section{Introduction}
In recent years the introduction of LLMs including \texttt{GPT} \citep{brown2020language, openai2024gpt4} and \texttt{Llama} \citep{touvron2023llamaopenefficientfoundation, touvron2023llama2openfoundation} has led to a significant advancement in NLP. However, expanding the reach of NLP to low-resource languages is crucial for advancing multilingual AI systems and promoting technological inclusivity. Urdu, with over 70 million native speakers, stands as a significant yet underserved language in the NLP domain \citep{blasi-etal-2022-systematic}.

For this study, we classify LLMs into two distinct categories:
\begin{enumerate}
    \item \textbf{Generalists}: General-purpose models capable of performing a wide variety of tasks. We will use \texttt{GPT-4-Turbo} (abbreviated as \texttt{GPT}) trained on dataset up to Dec 2023 and \texttt{Llama-3-8b} (abbreviated as \texttt{Llama}) as the generalist models.
    
    \item \textbf{Specialists}: Special-purpose models fine-tuned to perform specific tasks. We use \texttt{XLM-Roberta-large} (abbreviated as \texttt{XLM-R}) \citep{conneau-etal-2020-unsupervised}, \texttt{mT5-large} (abbreviated as \texttt{mT5}) \citep{xue-etal-2021-mt5}, and a fine-tuned version of \texttt{Llama-3-8b} (abbreviated as \texttt{Llama-FT}) as the specialist models.

\end{enumerate}

We present a comprehensive evaluation of generalist and specialist models for classification and generation tasks, exploring their strengths and limitations. Table \ref{table:all-tasks} outlines the sub-tasks associated with both categories, illustrating the scope of our evaluation.

\renewcommand{\arraystretch}{1.4}
\begin{table}[!ht]
    \small
    \begin{center}    
    \begin{tabularx}{\columnwidth}{|X|X|}
        \hline
        \textbf{Classification} & \textbf{Generation} \\
        \hline

        Sentiment Analysis & Question Answering \\
        \hline

        Abuse Detection & Summarization \\
        \hline

        Sarcasm Detection & Paraphrasing \\
        \hline

        Fake News Detection & Transliteration \\
        \hline

        Topic Classification & Translation (en-ur) \\
        \hline

        Part-of-Speech Tagging & Translation (ur-en) \\
        \hline

        Named-entity Recognition &  AI Assistant \\
        \hline
    \end{tabularx}
    \caption{\small{Sub-tasks for classification and generation.}}
    \label{table:all-tasks}
    \end{center}
    \vskip -0.1in
\end{table}

In this paper, we aim to answer: (1) how each category of models performs on Urdu language tasks, and (2) which model type is more effective in practical applications for Urdu-speaking users. Specifically, we seek to determine if the added specialization of the specialist models translates into significant performance gains over the generalist models in the defined tasks. 

Our contributions can be summarized as follows:
\begin{enumerate}
    \item We fine-tune \texttt{XLM-R}, \texttt{mT5}, and \texttt{Llama} on classification and generation tasks to optimize their performance for the defined tasks.
    \item We compare the performance of both generalist and specialist models on a smaller, controlled test set consisting of \texttt{min(1000, len(dataset["test"]))} samples, and provide a detailed comparison using various metrics. The exact size of each test set is given in Appendix \ref{sec:appendix-dataset}. Given the inherent challenges of working with low-resource languages like Urdu, our test set size reflects the current limitations in available data.
    \item We conduct a human evaluation of the outputs from the generation tasks and compare these results with the automated evaluations performed by \texttt{GPT}, \texttt{Llama} and \texttt{Claude 3.5 Sonnet} (abbreviated as \texttt{Claude}).
\end{enumerate}

The code, model outputs, and the human and LLM evaluations are publicly available on GitHub\footnote{\url{https://github.com/sameearif/Generalists-vs-Specialists}}.

\section{Related Work}

In recent years there has been a growing interest in the performance of LLMs across various languages. The MEGA benchmark \citep{ahuja2023mega} evaluates 16 NLP datasets across 70 languages. They compare the performance of \texttt{BLOOMZ}, \texttt{GPT} models, and State of the Art (SOTA) non-autoregressive models. MEGAVERSE \cite{ahuja2024megaverse} builds on top of the MEGA benchmark and evaluates the non-English capabilities of \texttt{GPT-3.5-Turbo}, \texttt{GPT-4}, \texttt{PaLM2}, \texttt{Gemini-Pro}, \texttt{Mistral}, \texttt{Llama-2}, and \texttt{Gemma}. IndicGenBench \citep{singh2024indicgenbench} evaluates LLMs on user-facing generation tasks across a set of 29 Indic languages. They perform evaluation on both proprietary and open-source LLMs including \texttt{GP-3.5}, \texttt{GPT-4}, \texttt{PaLM-2}, \texttt{mT5}, \texttt{Gemma}, \texttt{BLOOM}, and \texttt{Llama}. They cover the following tasks: cross-lingual summarization, machine translation, and cross-lingual question-answering. \citet{zhao2024llama} conduct an evaluation of \texttt{Llama}'s response quality based on LLM-Eval \citep{zhang2023llmeval}, a benchmark comprising instruction tasks from 17 categories. \citet{mujadia2024large} presents a comprehensive translation evaluation using \texttt{Llama} for English and Indian languages. They found that LLM-based evaluator achieves a comparable score with human judgement.

\citet{khondaker2023gptaraeval} presents a in-depth evaluaton of \texttt{BLOOMZ}, \texttt{ChatGPT} and specialist models \texttt{AraT5} and \texttt{MARBERTv2}. They evaluated these models for Arabic on 44 distinct language understanding and generation tasks. They also present a comparison between the human evaluation and \texttt{GPT-4} evaluation. Additionally, they observed that the English prompt works better than the Arabic prompt. \citet{abdelali2024larabench} provides a benchmark of LLMs against SOTA models for Arabic NLP. They evaluated \texttt{GPT-3.5-Turbo}, \texttt{GPT-4}, \texttt{BLOOMZ}, and \texttt{Jais-13b-Chat} across 33 tasks using 61 publicly available datasets, resulting in 330+ experimental setups. They observed that SOTA models generally outperform LLMs in zero-shot learning, larger models with few-shot learning techniques significantly reduce the performance gaps.

Existing multilingual NLP benchmarks often lack extensive language-specific evaluation and comparison against task-specific models. \citet{tahir2024benchmarking} address this gap by evaluating \texttt{GPT-3.5-Turbo}, \texttt{Llama-2-7B-Chat}, and \texttt{Bloomz (3B and 7B)} across 14 tasks using 15 Urdu datasets in a zero-shot setting, comparing their performance against task-specific models. Their findings reveal that task-specific models (Support Vector Machine (SVM), Decision Tree, and \texttt{m-BERT}, etc) generally outperform the mentioned LLMs in Urdu NLP tasks with zero-shot learning. However, they do not perform few-shot prompting, chain-of-thought prompting, or LLM fine-tuning.

\section{Datasets}
\subsection{Classification}

\textbf{Sentiment Analysis.} We use the Urdu IMDB sentiment analysis dataset\footnote{\url{https://github.com/urduhack/resources/releases/tag/imdb_urdu_reviews_v1.0.0}}, which is a translated version of the original IMDB dataset \citep{maas-EtAl:2011:ACL-HLT2011}. It is translated using Google Translator, and comprises 50,000 movie reviews. \\

\noindent \textbf{Abuse Detection.} We use the dataset by \citet{9094176} which has 2,171 entries and the dataset by \citet{amjad2022overview} which consists of 3,502 entries. \\ 

\noindent \textbf{Sarcasm Detection.} For this task we use Urdu Sarcastic Tweets Dataset \citep{shumaila_khan_fahad_najeeb_2023} which consists of 19,955 tagged tweets. \\

\noindent \textbf{Fake News Detection.} We use the fake news dataset by \citet{amjad-etal-2020-data} which is a mixture of real and translated data. It comprises 1,300 labeled news articles.  \\

\noindent \textbf{Topic Classification.} For this task, we use a dataset\footnote{\url{https://github.com/mwaseemrandhawa/Urdu-News-Headline-Dataset}} consisting of 137,161 news headlines categorized into the following topics: business, entertainment, health, politics, science, sports, world and other. \\

\begin{figure}[h!]
\begin{center}
\centerline{\includegraphics[width=1\columnwidth]{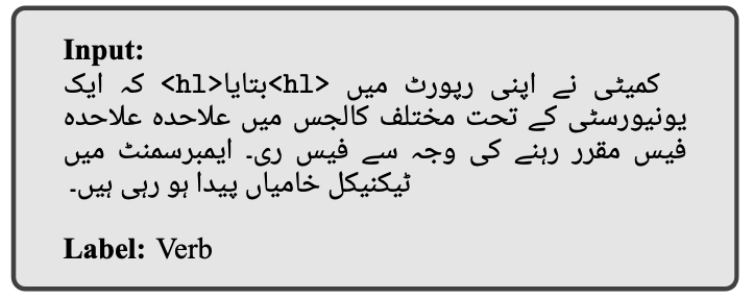}}
\caption{\small{PoS Data Structure for Llama}}
\label{fig:pos-struct}
\end{center}
\vskip -0.3in
\end{figure}

\noindent \textbf{Part-of-Speech (PoS) Tagging.} For this task, we use the Universal Dependencies \citep{nivre-etal-2020-universal} dataset, which consists of 5,130 sentences, annotated with a PoS tag for every word. For \texttt{GPT} and \texttt{Llama}, we wrap the word for which we want to predict the PoS tag in \texttt{<hl>} tags. The structure of the data is given in Figure \ref{fig:pos-struct}.

\noindent \textbf{Named Entity Recognition (NER).} We use the dataset\footnote{\url{https://huggingface.co/datasets/mirfan899/urdu-ner}} available on Hugging Face which consists of 33,748 sentences, annotated with a NER tag for every word. For \texttt{GPT} and \texttt{Llama}, we applied the same data structuring method as done in the PoS tagging task.

\subsection{Generation}
Given our resource constraints, we could not set the maximum sequence length for model training to the longest entries in each dataset. Instead, we filter the datasets to include only those entries that fell within a manageable maximum length of 2048. This approach allowed us to optimize the training process and ensure efficient use of computational resources while still maintaining a representative sample of the data. \\

\noindent \textbf{Question-Answering.} We use three datasets for question-answering: (1) UQA \citep{arif-etal-2024-uqa-corpus}, consisting of 88,829 answerable questions; (2) UQuAD\footnote{\url{https://github.com/ahsanfarooqui/UQuAD---Urdu-Question-Answer-Dataset/tree/main}}, containing 139 questions; and (3) Wiki-UQA\footnote{\url{https://huggingface.co/datasets/uqa/Wiki-UQA}}, a manually generated dataset from Wikipedia articles, comprising 210 questions. \\

\noindent \textbf{Summarization.} For this task, we use the XSUMUrdu \citep{munaf2023low} dataset, selecting a subset of 76,626 entries based on the maximum length used during model training. \\ 

\noindent \textbf{Paraphrasing.} For paraphrasing we use the dataset\footnote{\url{https://huggingface.co/datasets/mwz/ur_para}} available on Hugging Face. We select 387,004 entries from the dataset based on the maximum length used while training the models. \\ 

\noindent \textbf{Transliteration.} We use the Dakshina dataset \citep{roark-etal-2020-processing} from which we select 11,464 sentences based on the maximum length used while training the models. \\

\noindent\textbf{Translation.} We use OPUS-100 \citep{zhang-etal-2020-improving} \citep{tiedemann-2012-parallel} for English-to-Urdu and Urdu-to-English translation. We select 755,526 sentences from this dataset based on the maximum length used while training the models. \\

\noindent\textbf{AI Assistant.} We use UrduAssistant \citep{urduassistant} dataset which is translated to Urdu from English and has 67,017 prompts.

\section{Methodology}
% \subsection{Test Dataset and Metrics}
\subsection{Experimental Design}
We utilize a controlled test set consisting of \texttt{min(1000, len(dataset["test"]))} samples for each task except AI assistant, ensuring that the evaluation is both comprehensive and cost-efficient. For AI assistant we do human evaluation of 50 samples. We ensure that the test dataset is as balanced as possible, with an effort to achieve equal representation of different classes within each task. We use Macro-$F_1$ Score to evaluate all the classification tasks, SQuAD $F_1$ \citep{rajpurkar-etal-2018-know} \cite{rajpurkar-etal-2016-squad} to evaluate question-answering, SacreBLEU \citep{post2018clarity} for paraphrasing, transliteration and translation, ROUGE-L (with word-level tokenization) \cite{lin-2004-rouge} for summarization and Wins (the number of times a model is ranked one by the evaluator) for AI assistant evaluation.

% \subsection{Specialist Models' Fine-tuning}
We fine-tune \texttt{Llama} and \texttt{XLM-R} for each classification task separately, ensuring that each model is specifically optimized for its respective task.  Similarly, for generation tasks, we fine-tune the \texttt{mT5} and \texttt{Llama} separately for each task. For the \texttt{mT5} models, we use a learning rate of $5e^{-5}$. The \texttt{Llama} models are fine-tuned with a learning rate of $2e^{-4}$ for both generation and classification tasks. In the case of \texttt{XLM-R}, we use a learning rate of $5e^{-6}$. We use LoRA \citep{hu2021lora} to fine-tune int4 quantized \texttt{Llama}. The batch size, number of epochs each model is trained for and LoRA config for \texttt{Llama} is given in Appendix \ref{sec:appendix-hyperparam}. After fine-tuning all the models we perform evaluation on the test dataset.

\begin{figure}[t]
\begin{center}
\centerline{\includegraphics[width=1\columnwidth]{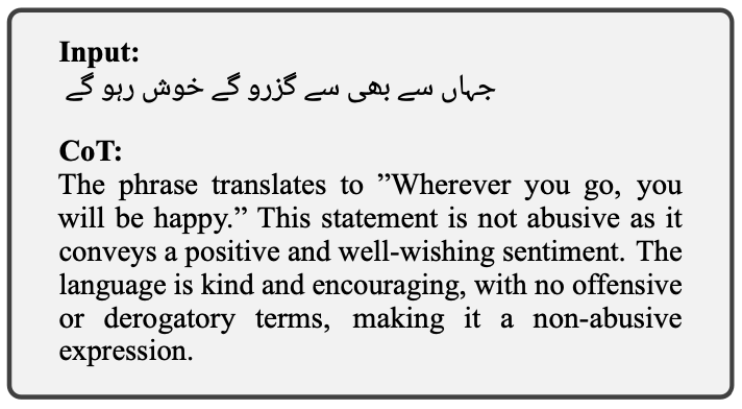}}
\caption{\small{CoT example from abuse detection}}
\label{fig:abusedetection}
\end{center}
\vskip -0.2in
\end{figure}

% \subsection{Generalist Models}
To evaluate the performance of the generalist models, we design a series of experiments. We use \texttt{GPT} and \texttt{Llama} as our generalist models. \texttt{GPT} is chosen due its top performance on the LMSYS chatbot arena \citep{chiang2024chatbot} as of March 1st, 2024 when we started our research. Our experimental setup is as follows:
We conduct experiments under zero-shot, three-shot, and six-shot settings for both generation and classification tasks. The examples for the three-shot and six-shot scenarios are selected from the training dataset of each task. Specifically, the examples are carefully selected by a human expert to ensure a representative and balanced sample, avoiding any unintended bias in the selection process. We also conduct experiments with Chain-of-Thought (CoT) reasoning for classification tasks in the six-shot setting. We create CoT reasoning for the six selected examples from the training dataset. These examples along with their CoT is given as a few-shot prompt to the model. Figure \ref{fig:abusedetection} presents an example of generated CoT reasoning. For all the evaluations, the temperature and nucleus sampling for \texttt{GPT-4} is set to 1.0, which is the default for the \texttt{GPT} API. For \texttt{Llama}, the temperature is set to 0.6 and nucleus sampling is set to 0.9, i.e., the values used in the original code-base for \texttt{Llama}\footnote{\url{https://github.com/meta-llama/llama3/blob/main/example_chat_completion.py}}.

\subsection{Prompt Design}
\citet{khondaker-etal-2023-gptaraeval} observe in their \texttt{ChatGPT} evaluation for Arabic that an English prompt performs better than the Arabic one. Therefore, following this study, we decide to use English prompts for our evaluations of Urdu tasks as well. In the prompt templates, the following placeholders are used:
\begin{enumerate}
    \item \textcolor{blue}{ROLE}: It specifies the persona of the LLM. For example, it could be sentiment classifier, abuse detector, sarcasm detector, etc.
    \item \textcolor{orange}{TASK DESCRIPTION}: It provides a brief description of what the task is and what the model is expected to do.
    \item \textcolor{red}{LABEL LIST}: It lists the possible labels the model can assign to the input text. For example, it could be \texttt{['positive', 'negative']} for sentiment analysis.
\end{enumerate}

\begin{figure}[h!]
\begin{center}
\centerline{\includegraphics[width=1\columnwidth]{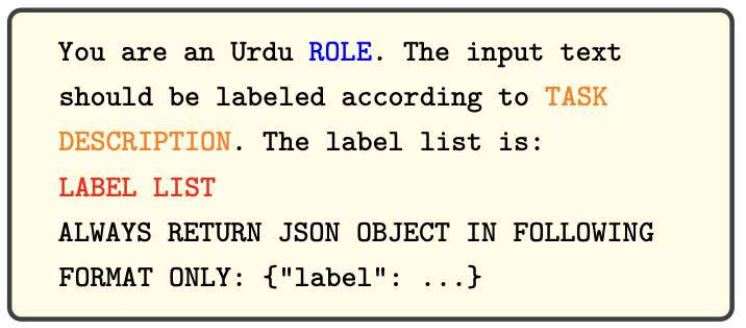}}
\caption{\small{Classification Prompt Template}}
\label{fig:classificationtemplate}
\end{center}
\vskip -0.3in
\end{figure}

\begin{figure}[h!]
\begin{center}
\centerline{\includegraphics[width=1\columnwidth]{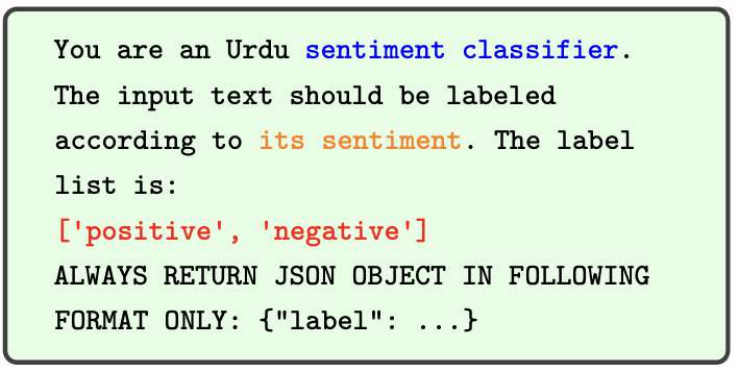}}
\caption{\small{Classification Prompt Example}}
\label{fig:classificationexample}
\end{center}
\vskip -0.3in
\end{figure}

\begin{figure}[t]
\begin{center}
\centerline{\includegraphics[width=1\columnwidth]{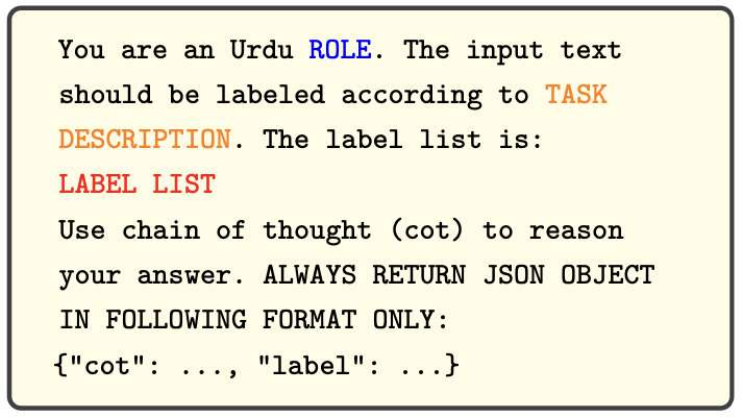}}
\caption{\small{Classification Prompt Template (CoT)}}
\label{fig:classificationcottemplate}
\end{center}
\vskip -0.3in
\end{figure}

\begin{figure}[t]
\begin{center}
\centerline{\includegraphics[width=1\columnwidth]{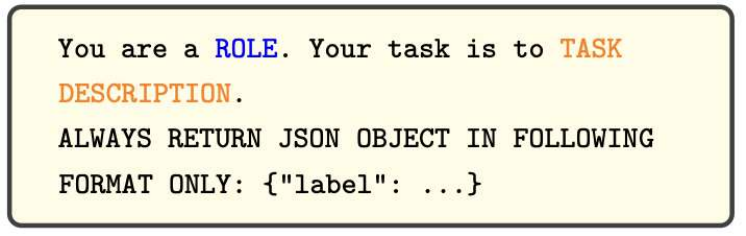}}
\caption{\small{Generation Prompt Template}}
\label{fig:generationtemplate}
\end{center}
\vskip -0.3in
\end{figure}

\begin{figure}[t]
\begin{center}
\centerline{\includegraphics[width=1\columnwidth]{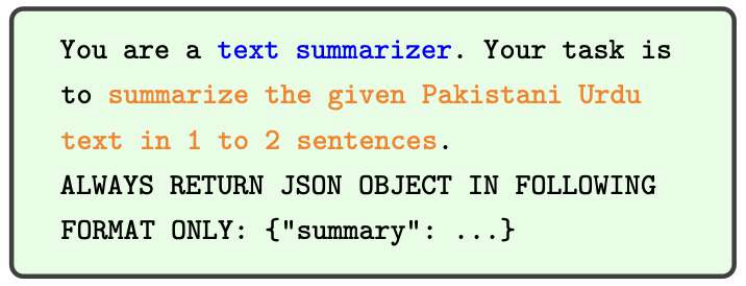}}
\caption{\small{Generation Prompt Example}}
\label{fig:generationexample}
\end{center}
\vskip -0.3in
\end{figure}

Figure \ref{fig:classificationtemplate} shows the prompt template for the classification task without CoT, and Figure \ref{fig:classificationexample} shows an example prompt for it. Figure \ref{fig:classificationcottemplate} shows the CoT prompt template for the classification task. Figure \ref{fig:generationtemplate} presents the prompt template for the generation task, and Figure \ref{fig:generationexample} provides an example of it. Appendix \ref{sec:appendix-prompt-criteria} contains all the prompts for classification and generation tasks.

\subsection{Human Evaluation}
We select a subset of size 50 (as done by \citet{khondaker2023gptaraeval} for Arabic) from the test dataset of each task. For human evaluation, two annotators (native Urdu speakers) are presented with anonymized outputs of \texttt{Llama}, \texttt{Llama-FT}, \texttt{mT5} and \texttt{GPT}. They are asked to rank them from one to four based on the criteria in Appendix \ref{sec:appendix-prompt-criteria}, allowing multiple models to have the same rank. We prompt \texttt{GPT}, \texttt{Llama} and \texttt{Claude} with the same criteria, to assign a score to the outputs of the mentioned tasks, and then ranking is done based on this score. Since both \texttt{GPT} and \texttt{Llama} responses are being evaluated, we also include \texttt{Claude} as an additional evaluator because of the bias when LLMs assess their own outputs \citep{arif2024fellowshipllmsmultiagentworkflows}. We compare ranking done by human annotators, and the three LLMs using Krippendorff's alpha to determine the inter-rater reliability to determine the determine whether LLMs are effective evaluators for Urdu tasks.

\section{Evaluation and Discussion}
\subsection{Classification}
\begin{center}
\begin{table*}[ht!]
\centering
\renewcommand{\arraystretch}{1}
\small % Change font size to small
\begin{tabular}{lccccccccccccc}
\toprule
& \multicolumn{4}{c}{\texttt{GPT}} & \multicolumn{4}{c}{\texttt{Llama}} & \texttt{XLM-R} & \texttt{Llama-FT} \\
\cmidrule(lr){2-5}
\cmidrule(lr){6-9}
Task & 0 & 3 & 6 & CoT & 0 & 3 & 6 & CoT  \\
\midrule
Sentiment Analysis & 90.98 & 91.17 & 94.90 & 94.60 & 87.88 & 60.62 & 59.71 & 57.54 & 92.90 & \textbf{95.30} \\
Abuse Detection & 86.27 & 89.01 & 88.71 & 87.62 & 44.73 & 65.85 & 71.64 & 75.82 & 
\textbf{90.92} & 89.15 \\
Sarcasm Detection & 58.17 & 69.18 & 66.56 & 65.47 & 44.94 & 34.25 & 50.67 & 49.75 & \textbf{84.37}  & 81.48 \\
Fake News Detection & 78.88 & 75.95 & 78.14 & 76.45 & 66.36 & 63.48 & 71.45 & 40.98 & \textbf{84.99} & 72.14 \\
Topic Classification & 76.05 & 73.54 & 74.57 & 70.43 & 54.08 & 53.02 & 53.68 & 49.24 & 84.49 & \textbf{84.74} \\
PoS Tagging & 53.31 & 51.51 & 54.17 & 54.61 & 25.80 & 34.32 & 33.85 & 41.62 & 65.41 & \textbf{67.55} \\
NER Tagging & 61.96 & 62.18 & 62.98 & 63.95 & 42.13 & 40.80 & 45.29 & 53.96 & 70.41 & \textbf{90.90} \\
\bottomrule
\end{tabular}
\caption{\small{We report Macro-$F_1$ score for each classification task. \texttt{GPT-4-Turbo} and \texttt{Llama-3-8b} is evaluated in 0-shot, 3-shot, 6-shot and 6-shot with Chain-of-Thought settings. FT stands for fine-tuned.}}
\label{tab:classificationresults}
\end{table*}
\vskip -0.2in
\end{center}

We present the evaluation of the generalists (\texttt{GPT}, \texttt{Llama}) and the specialists (\texttt{Llama-FT}, \texttt{XLM-R}) for the classification tasks in Table \ref{tab:classificationresults}. We observe that \texttt{Llama-FT} (fine-tuned) model achieves the highest scores in four tasks (sentiment analysis, topic classification, PoS tagging, and NER tagging), while \texttt{XLM-R} outperforms other models in three tasks (abuse detection, sarcasm detection, and fake news detection). \texttt{GPT} does not perform better than \texttt{Llama-FT} and \texttt{XLM-R}. For certain tasks like NER and PoS tagging, CoT reasoning leads to a better Macro-$F_1$ score. We now discuss the performance on each classification task in detail. \\

\noindent \textbf{Sentiment Analysis.} For sentiment analysis, \texttt{Llama-FT} achieves the highest score with a Macro-$F_1$ of 95.30, outperforming \texttt{XLM-R} and \texttt{GPT}. \texttt{GPT}'s performance improves with more shots, reaching a Macro-$F_1$ of 94.90 with 6-shot learning. The CoT setting provides a slight increase to 94.60. However the Macro-$F_1$ for \texttt{Llama} decreases with few-shot and CoT prompting. The small difference in the Macro-$F_1$ between \texttt{Llama-FT} and \texttt{GPT} indicates the effectiveness of \texttt{GPT} for Urdu sentiment analysis without requiring task-specific fine-tuning.  \\

\begin{figure}[t]
\begin{center}
\centerline{\includegraphics[height=4.6in]{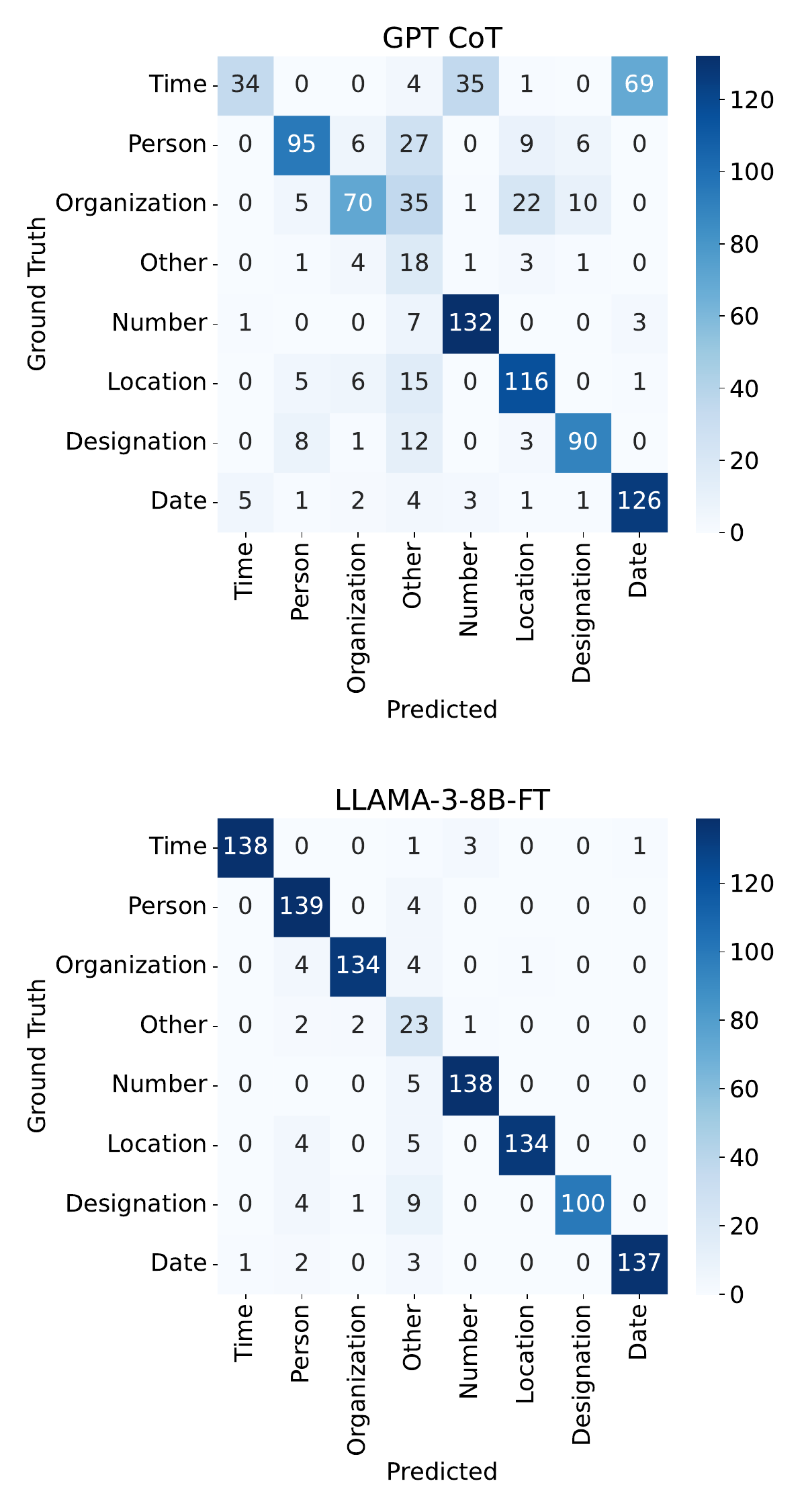}}
\caption{\small{Performance of \texttt{GPT-CoT} and \texttt{Llama-FT} on recognizing different entities in the NER task.}}
\label{fig:nertaggingcomparison}
\end{center}
\vskip -0.3in
\end{figure}

\noindent \textbf{NER Tagging.} For NER tagging, \texttt{Llama-FT} achieves the highest score with a Macro-$F_1$ of 90.90, significantly outperforming \texttt{XLM-R} and \texttt{GPT}. The highest score achieved by \texttt{GPT} is 63.95, and by \texttt{Llama} is 53.96, both using CoT reasoning. We uncover the differences in accuracy for various entities for \texttt{GPT} and \texttt{Llama-FT} in Figure \ref{fig:nertaggingcomparison}. We observe a drop in the recognition accuracy of Person entities by \texttt{GPT}, with 95 correctly recognized compared to 139 by \texttt{Llama-FT}. We notice a similar trend for Organization entities, with 70 classified correctly compared to 134 by \texttt{Llama-FT}. This suggests that NER fine-tuning on a specialized Urdu dataset improves the model's ability to recognize Person and Organization entities in Urdu text. \\

\noindent \textbf{Abuse Detection.} In the task of abuse detection, \texttt{XLM-R} leads with a Macro-$F_1$ of 90.92, followed closely by \texttt{Llama-FT} at 89.15. \texttt{GPT} achieves its highest performance of 89.01 with 3-shot prompting, while \texttt{Llama} reaches a peak of 75.82 using CoT reasoning. \\

\noindent \textbf{Sarcasm Detection.} \texttt{XLM-R} performs best in sarcasm detection with a Macro-$F_1$ score of 84.37. \texttt{Llama-FT} also shows strong performance with 81.48. \texttt{GPT} achieves its best performance at 69.18 with 3-shot prompting, while \texttt{Llama} performs best at 50.67 with 6-shot prompting. The large difference between the Macro-$F_1$ scores of generalist and specialist models indicates that sarcasm detection in Urdu can be challenging without task-specific fine-tuning. \\

\noindent \textbf{Fake News Detection.} For fake news detection, \texttt{XLM-R} achieves the highest score with a Macro-$F_1$ of 84.99, while \texttt{Llama-FT} follows with 72.14. \texttt{GPT}'s performance is relatively consistent across different shot settings, with its highest score being 78.88 in the 0-shot setting. \texttt{Llama}'s highest score is 71.45 at 6-shot setting. The large margin by \texttt{XLM-R} indicates its effectiveness in discerning fake news.

\begin{figure}[t]
\begin{center}
\centerline{\includegraphics[width=0.9\columnwidth]{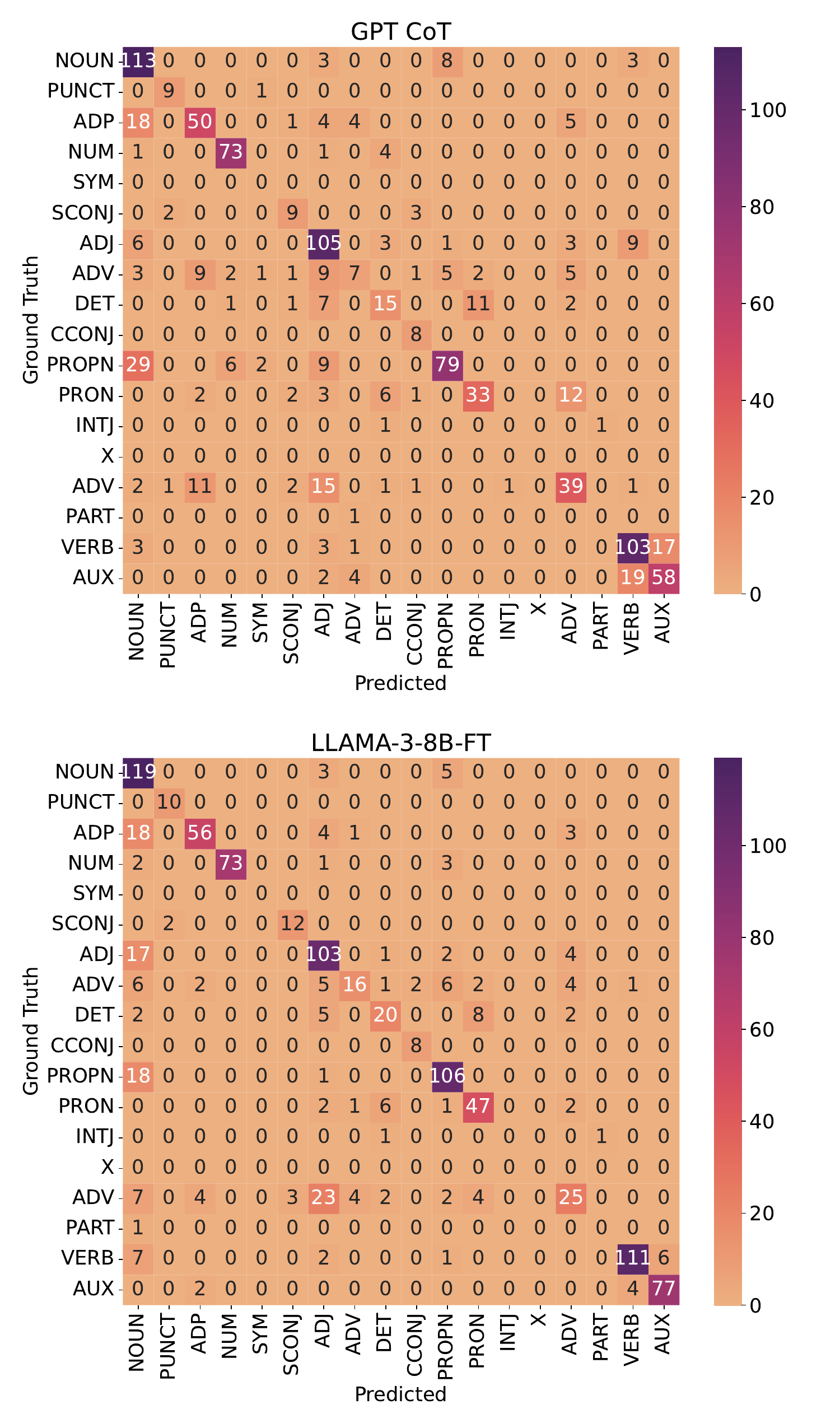}}
\caption{\small{Performance of \texttt{GPT-CoT} and \texttt{Llama-FT} on recognizing different parts of speech in the POS task.}}
\label{fig:postaggingcomparison}
\end{center}
\vskip -0.35in
\end{figure}

\begin{center}
\begin{table*}[ht]
\centering
\renewcommand{\arraystretch}{1}
\small
\begin{tabular}{llcccccccc}
\toprule
&  & \multicolumn{3}{c}{\texttt{GPT}} & \multicolumn{3}{c}{\texttt{Llama}} & \texttt{mT5} & \texttt{Llama-FT} \\
\cmidrule(lr){3-5}
\cmidrule(lr){6-8}
Task & Metric & 0 & 3 & 6 & 0 & 3 & 6  \\
\midrule
Question-Answering & SQuAD-$F_1$ & 66.28 & 72.55 & \textbf{72.78} & 69.89 & 71.65 & 71.62 & 69.66 & 70.42 \\
Summarization & ROUGE-L & 22.54 & 23.35 & 23.34 & 25.76 & 25.84 & 26.17 & 30.72 & \textbf{31.01} \\
Paraphrasing & SacreBLEU & 3.59 & 3.92 & 4.01 & 2.42 & 6.40 & 6.11 & \textbf{11.98} & 10.17 \\
Transliteration & SacreBLEU & 30.93 & 32.38 & 32.09 & 12.37 & 18.51 & 21.35 & \textbf{40.23} & 37.95 \\
Translation (en-ur) & SacreBLEU & 12.07 & 12.62 & 11.59 & 5.14 & 6.56 & 5.83 & \textbf{18.35} & 14.63 \\
Translation (ur-en) & SacreBLEU & 16.29 & 18.05 & 19.18 & 7.83 & 13.11 & 12.32 & 21.55 & \textbf{29.18} \\
\bottomrule
\end{tabular}
\caption{\small{We report SQuAD-$F_1$, ROUGE-L or SacreBLEU depending on the generation tasks. \texttt{GPT-4-Turbo} and \texttt{Llama-3-8b} is evaluated in 0-shot, 3-shot and 6-shot setting.}}
\label{tab:generationresults}
\end{table*}
\end{center}
\vskip -0.17in

\noindent \textbf{Topic Classification.} \texttt{Llama-FT} excels in topic classification with a Macro-$F_1$ score of 84.74, significantly higher than the other models. \texttt{XLM-R} also performs well with 84.49. \\

\noindent \textbf{PoS Tagging.} In PoS tagging, \texttt{Llama-FT} leads with a Macro-$F_1$ of 67.55, followed by \texttt{XLM-R} at 65.41. \texttt{GPT}'s highest score is 53.31 in the 0-shot setting and \texttt{Llama}'s highest score is 41.62 with CoT reasoning. To understand the significant difference in Macro-$F_1$ between \texttt{GPT} and \texttt{Llama-FT}, we study the individual class performance (Figure \ref{fig:postaggingcomparison}). We find that \texttt{GPT} struggles with correctly tagging proper nouns, pronouns, and auxiliaries, while \texttt{Llama-FT} is able to identify most of them correctly, suggesting that task-specific fine-tuning improves tagging performance for these parts of speech.

\subsection{Generation}
We present the evaluation of the generalists and the specialists for the generation tasks in Table \ref{tab:generationresults} and Table \ref{tab:win-rates-all}. We observe that the fine-tuned \texttt{Llama-FT} model achieves the highest scores in summarization and translation from Urdu to English. On the other hand \texttt{GPT} shows a consistent performance across all the tasks with its best results often appearing in the 6-shot setting. The \texttt{mT5} model also demonstrates strong performance in tasks such as transliteration and paraphrasing, benefiting from its extensive multilingual training on translation tasks. We now discuss the performance on each generation task in detail.

\begin{center}
\begin{table*}[ht]
\centering
\renewcommand{\arraystretch}{1}
\small % Change font size to small
\begin{tabular}{lcccccccc} % Add correct number of columns here
\toprule
& \multicolumn{2}{c}{\texttt{GPT}'s Wins} & \multicolumn{2}{c}{\texttt{Llama}'s Wins} & \multicolumn{2}{c}{\texttt{mT5}'s Wins} & \multicolumn{2}{c}{\texttt{Llama-FT}'s Wins} \\
\cmidrule(lr){2-3}
\cmidrule(lr){4-5}
\cmidrule(lr){6-7}
\cmidrule(lr){8-9}
Task & Annot. 1 & Annot. 2 & Annot. 1 & Annot. 2 & Annot. 1 & Annot. 2 & Annot. 1 & Annot. 2 \\
\midrule
Summarization & 41 & 34 & 9 & 17 & 7 & 12 & 4 & 11 \\
Paraphrasing & 42 & 39 & 12 & 15 & 4 & 3 & 4 & 4 \\
Transliteration & 40 & 38 & 5 & 6 & 13 & 21 & 19 & 21 \\
Translation (en-ur) & 46 & 46 & 3 & 1 & 7 & 10 & 13 & 10 \\
Translation (ur-en) & 40 & 45 & 12 & 10 & 14 & 15 & 18 & 15 \\
AI Assistant & 49 & 49 & 3 & 3 & 0 & 0 & 1 & 1 \\
\bottomrule
\end{tabular}
\caption{\small{We report the number of times each model is  ranked 1 by each annotator for the generation tasks.}}
\label{tab:win-rates-all}
\end{table*}
\end{center}

\subsubsection{Quantitative Evaluation}
\noindent \textbf{Question-Answering.} All the models show similar performance on question-answering task. \texttt{GPT} achieves the highest score with a SQuAD-F1 of 72.78 in the 6-shot setting. \texttt{Llama} closely follows with a score of 71.65 in 3-shot setting. \texttt{Llama-FT} has a lower score of 70.42, indicating that the translated UQA dataset is not very effective in improving the performance of the model. \texttt{mT5} also performs well with a score of 69.66. The consistent improvement of \texttt{GPT} with increasing shots suggests its capability to use more context effectively. \\

\noindent \textbf{Summarization.} In the summarization task, \texttt{Llama-FT} achieves the highest ROUGE-L score of 31.01, outperforming \texttt{Llama} and \texttt{GPT}. \texttt{mT5} also shows strong performance with a score of 30.72. Non-fine-tuned \texttt{Llama} outperforms \texttt{GPT} in all few-shot settings. \texttt{GPT}’s best performance is in the 3-shot setting with a score of 23.35 while \texttt{Llama}'s best is at 6-shot setting with a score of 26.17. The fine-tuning of \texttt{Llama} contributes to its superior performance in capturing and summarizing Urdu content effectively. \\

\noindent \textbf{Paraphrasing.} For paraphrasing, \texttt{mT5} achieves the highest SacreBLEU score of 11.98. This indicates the advantage \texttt{mT5} has due to its massive multilingual pre-training. \texttt{Llama-FT} follows with a SacreBLEU score of 10.17. \texttt{Llama}'s best score is 6.40 at 3-shot setting outperforming \texttt{GPT}’s best score of 4.01. \\ 

\begin{figure}[htbp]
\begin{center}
\centerline{\includegraphics[width=\columnwidth]{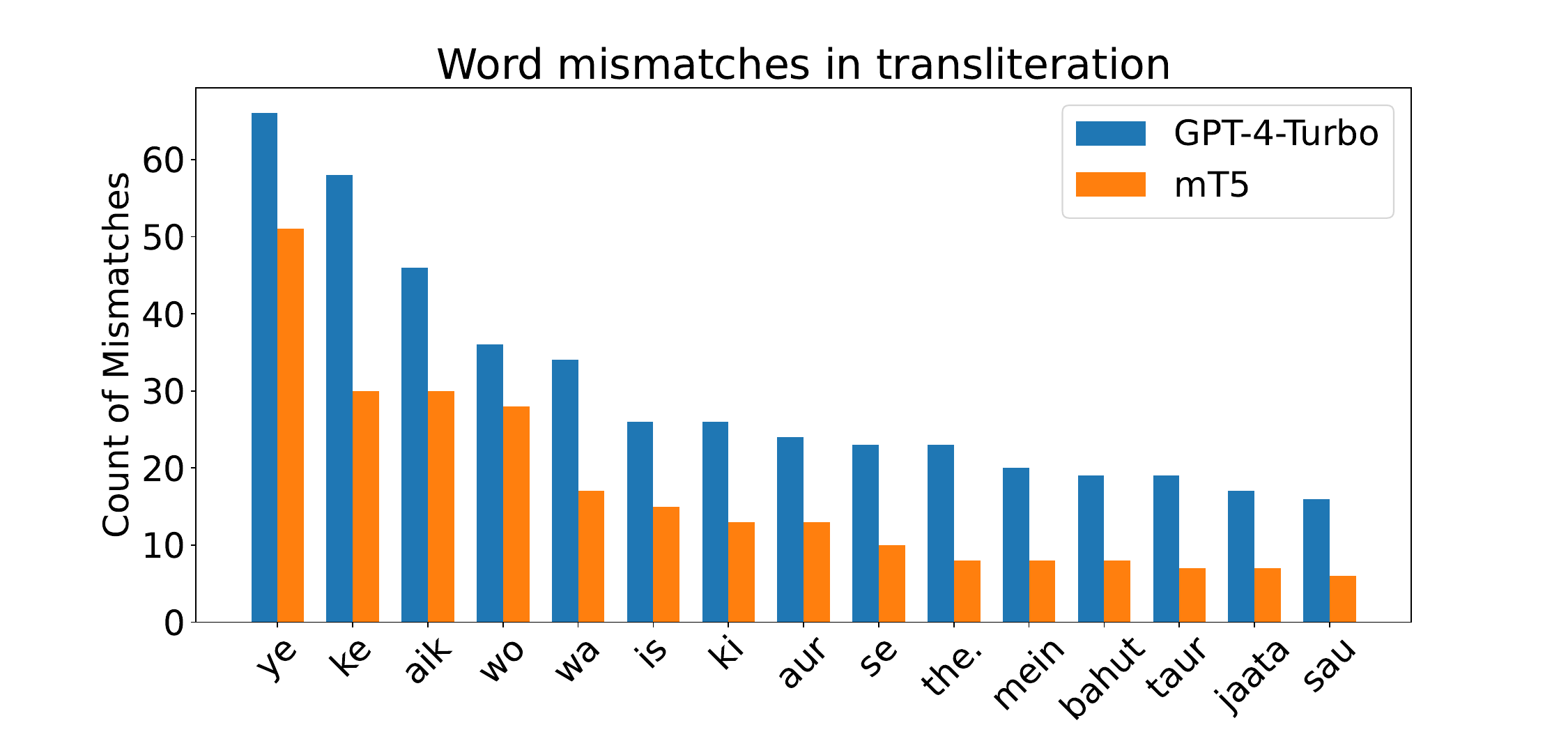}}
\caption{\small{Comparison between mismatches in the words predicted by \texttt{mT5} and \texttt{GPT} for the transliteration task.}}
\label{fig:transliterationcomparison}
\end{center}
\vskip -0.2in
\end{figure}

\noindent \textbf{Transliteration.} In the task of transliteration, \texttt{mT5} leads with a SacreBLEU score of 40.23, followed by \texttt{Llama-FT} at 37.95. \texttt{GPT}’s performance peaks at 32.38 with 3-shot learning. \texttt{Llama} shows poor performace in this task with the highest score of 21.35 at 6-shot setting. Figure \ref{fig:transliterationcomparison} shows the words with the highest mismatches in the transliterated text. To count these mismatches, we first tokenize the transliterated sentences and then count the instances where the predicted word differs from the ground truth. Smaller words such as “ye,” “ke,” “aik,” and “wo” pose challenges for \texttt{GPT}, resulting in higher mismatch counts. In contrast, \texttt{mT5} demonstrates lower mismatches for these words. \\

\noindent \textbf{Translation (en-ur).} For English to Urdu translation, \texttt{mT5} achieves the highest SacreBLEU score of 18.35, indicating its proficiency in translating English to Urdu. \texttt{Llama-FT} follows with a score of 14.63. \texttt{GPT}’s performance is consistent, with its highest score being 12.62 in the 3-shot setting. \texttt{Llama} under-performs with a least highest score of 6.56 with 3-shot prompting. The superior performance of \texttt{mT5} is likely due to the inclusion of high-quality Urdu data in the multilingual C4 corpus used for its pre-training \cite{xue-etal-2021-mt5}. \\

\noindent \textbf{Translation (ur-en).} In Urdu to English translation, \texttt{Llama-FT} excels with a SacreBLEU score of 29.18, outperforming other models. Surprisingly, contrary to the results in en-ur translation, \texttt{mT5} shows a lower SacreBLEU of 21.55 compared to \texttt{Llama-FT}. \texttt{GPT}’s highest score is 19.18 in the 6-shot setting and \texttt{Llama}'s higest score is 13.11 in the 3-shot setting.

\subsubsection{Qualitative Evaluation}
\noindent \textbf{AI Assistant.} \texttt{GPT} outperforms all the models with 49 out of 50 wins as shown in Table \ref{tab:win-rates-all}. \texttt{Llama-FT} is ranked one 0 times according to both annotators and \texttt{mT5} is ranked number one only 1 time. The low performance of fine-tuned models, such as \texttt{Llama-FT} and \texttt{mT5}, can be attributed to the fact that the fine-tuning dataset was translated from English. \\

\begin{figure}[htbp]
\begin{center}
\centerline{\includegraphics[width=\columnwidth]{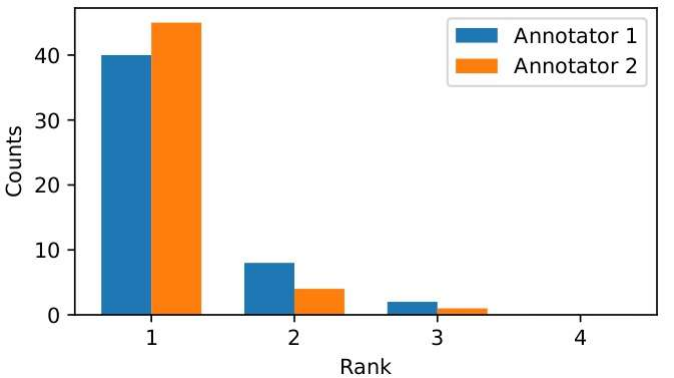}}
\small
\caption{\small{Rank counts for GPT in human evaluation of Urdu to English translation.}}
\label{fig:rankcomp}
\end{center}
\vskip -0.3in
\end{figure}

\noindent Based on quantitative evaluation, \texttt{GPT} only performs the best in the question-answering task based on the quantitative evaluation provided in Table \ref{tab:generationresults}. However, Table \ref{tab:win-rates-all} shows that \texttt{GPT} performs the best across all other tasks as well, according to qualitative evaluations by human annotators with more than 90\% win-rate for most of the tasks. Figure \ref{fig:rankcomp} presents the number of times each rank was assigned to \texttt{GPT} for Urdu to English translation. Appendix \ref{sec:appendix-human-eval} contains presents the rank counts for the other tasks.

Based on human evaluation \texttt{LLama} outperforms \texttt{mT5} and even the fine-tuned \texttt{Llama-FT} on summarization, paraphrasing and AI assistant task. On transliteration and both translation tasks \texttt{Llama-FT} outperforms \texttt{mT5}. This discrepancy between quantitative and qualitative evaluations highlights an important issue: the translation of datasets from other languages may introduce inconsistencies or artifacts that impact the quantitative metrics, especially for tasks like summarization, paraphrasing, and story generation. This observation stresses the need for the creation of native Urdu datasets, rather than relying on translations of existing datasets. It also reflects the limitations and lower quality of the currently available Urdu datasets, which further exacerbates these challenges. 

\subsection{Factors Contributing to Improved Performance}
Task-specific fine-tuning on Urdu datasets significantly improves the performance of LLMs on Urdu tasks, as evidenced by the superior results of fine-tuned \texttt{Llama}, \texttt{XLM-R}, and \texttt{mT5} on both classification and generation tasks. Additionally, the inclusion of high-quality Urdu data in pre-training datasets, such as the clean multilingual C4 corpus, enhances downstream performance, with \texttt{mT5} achieving higher SacreBLEU scores on transliteration, translation, and paraphrasing tasks compared to \texttt{GPT} and fine-tuned \texttt{Llama}. Providing more context in prompts also boosts performance, as shown by \texttt{GPT}'s improvement with increasing shots (0, 3, 6). Moreover, high-quality data for Indic languages like Hindi in pre-training improves performance on Urdu, especially for models like \texttt{XLM-R}, which excel in cross-lingual tasks, outperforming others in abuse detection, sarcasm detection, and fake news detection. The lack of native Urdu datasets poses a significant challenge for Urdu NLP, as most available datasets are translated from other languages.

\subsection{Human Evaluators vs. LLMs}
We compare human evaluation and LLM-based evaluation for summarization, paraphrasing, transliteration, English to Urdu translation, and Urdu to English translation, presented in Table \ref{tab:krippalpha}.

\begin{center}
\begin{table}[htbp]
\centering
\renewcommand{\arraystretch}{1}
\small % Change font size to small
\begin{tabular}{lcccc}
\toprule
Task & A & B & C & D \\
\midrule
Summarization & 0.684 & 0.504 & 0.502 & 0.624 \\
Paraphrasing & 0.710 & 0.592 & 0.471 & 0.565 \\
Transliteration & 0.694 & 0.510 & 0.253 & 0.603 \\
Translation (en-ur) & 0.728 & 0.592 & 0.392 & 0.515 \\
Translation (ur-en) & 0.730 & 0.474 & 0.307 & 0.510 \\
AI Assistant & 0.894 & 0.767 & 0.721 & 0.775 \\
\bottomrule
\end{tabular}
\caption{\small{\textbf{A}: Krippendorff's alpha between annotator 1 and annotator 2. \textbf{B}: Alpha between both annotators and \texttt{GPT}. \textbf{C}: Alpha between both annotators and \texttt{Llama}. \textbf{D}: Alpha between both annotators and \texttt{Claude}.}}
\label{tab:krippalpha}
\end{table}
\vskip -0.2in
\end{center}

For each task, the Krippendorff's alpha value for human evaluation exceeds 0.67, which, according to
Krippendorff’s interpretation is sufficient for a tentative conclusion to be drawn. Table \ref{tab:krippalpha} illustrates that the inclusion of \texttt{GPT}'s evaluation, \texttt{Llama}'s evaluation or \texttt{Claude}'s evaluation significantly reduces the alpha values meaning that the annotations done by \texttt{GTP}, \texttt{Llama} and \texttt{Claude} have a lower degree of agreement with the human annotators. For summarization, transliteration, Urdu-to-English translation, and AI assistant tasks, \texttt{Claude} has the highest agreement with humans, at 0.624, 0.604, 0.510, and 0.775, respectively. This showcases that the LLMs exhibit bias when the evaluated responses include their own generated content. For paraphrasing and Urdu-to-English translation tasks, \texttt{GPT} has the highest alpha values of 0.592 and 0.474, respectively. There is a higher agreement between the \texttt{GPT} rankings and human rankings as compared to the agreement between \texttt{Llama} rankings and human rankings.

\section{Conclusion}
In this paper, we present a comprehensive evaluation of generalist models and specialist models on 7 classification tasks and 7 generation tasks for Urdu NLP. Our evaluation covers prompting techniques such as few-shot, CoT reasoning as well as the fine-tuning of LLMs. We found that specialist models quantitatively outperformed generalist models on 12 out of the 14 tasks. The results highlight the importance of fine-tuning models to achieve higher performance in domain-specific applications in a low-resource setting. However, generalist models, such as \texttt{GPT}, showcased better performance in the human evaluation of the generation tasks, highlighting the importance of qualitative evaluation in accurately assessing model performance. We also performed a LLM based evaluation of the outputs of the models for the generation tasks. The low agreement between the rankings done by LLMs and human rankings shows that LLMs struggle when it comes to low-resource language understanding. Furthermore, the evaluation indicates a bias in LLMs when assessing responses that include their own outputs.

\section{Future Work}
One avenue for future research is to explore other strong general-purpose models (e.g., \texttt{GPT-4o}, \texttt{GPT-4o1} and \texttt{Claude}) and expand the scope of the evaluation to more Urdu NLP tasks. Additionally, using Retrieval-Augmented Generation (RAG) to find examples from training dataset for few-shot prompting would be an interesting experiment to enhance the performance of generalist models. In conclusion, while specialist models currently hold an edge in classification tasks performance, generalist models' adaptability and better performance in generation tasks remains valuable, and continuous advancements in LLMs promise further improvements for low-resource NLP.

\section{Limitations}
While our study provides valuable insights into the performance of generalist and specialist models for Urdu NLP tasks, it is important to acknowledge several limitations. The question-answering and sentiment analysis datasets used for training the specialist models are translated from English to Urdu. This translation process can introduce inaccuracies that may affect the models' performance. Additionally, the lack of native, high-quality Urdu datasets poses a significant challenge. Translated datasets often fail to capture the linguistic and cultural nuances inherent to Urdu, which can impact both training and evaluation outcomes. The evaluation is conducted on a subset of 1000 data points for each task. While this size is manageable and allows for a cost-efficient evaluation, it may not be fully representative of the model's overall performance. Although we use multiple annotators and calculate inter-rater reliability using Krippendorff's alpha, there is still a degree of subjectivity that may influence the results.

\section{Ethical Impact}
In this paper we present a comprehensive evaluation of LLMs with the aim to enhance the accessibility of NLP applications for Urdu speakers. This has significant ethical implications, as it addresses the digital divide and promotes linguistic diversity in technology. Our findings indicate that specialist models perform better than the generalist models in most Urdu NLP tasks. Consequently, our work may inspire researchers to develop more resources for the Urdu language, including models and datasets.

The potential risks associated with the usage of LLMs include the amplification of existing biases present in their training data, which may lead to unfair and discriminatory outcomes \citep{ye2023assessing}. A comprehensive fairness evaluation of these models must be conducted before they are deployed for public use.

\section*{Acknowledgments}
We are grateful for the time and effort put in by our research intern, Muhammad Suhaib Rashid who annotated our data and Mustafa Abbas who worked on creating Wiki-UQA dataset. We are also grateful to OpenAI for supporting our work through their Research Access Program.

\bibliography{acl_latex}

\appendix
\section{Implementation Details}
\subsection{Models}
\texttt{XLM-Roberta-large} is availabe on Hugging Face\footnote{\url{https://huggingface.co/FacebookAI/xlm-roberta-large}} under MIT license. \texttt{mT5-large} is available on Hugging Face\footnote{\url{https://huggingface.co/google/mt5-large}} under Apache-2.0 license. \texttt{Llama-3-8b} is available on Hugging Face\footnote{\url{https://huggingface.co/meta-llama/Meta-Llama-3-8B}} under llama3 license. \texttt{GPT-4} is available under 	proprietary licence. All models used in this paper comply with their respective license.

\subsection{Datasets}
Urdu IMDB sentiment analysis dataset OBDL license. Urdu NER dataset is availabe under MIT license. The abuse detection dataset by \citet{9094176} and by \citet{amjad2022overview}, Sarcastic Tweets Dataset \citep{shumaila_khan_fahad_najeeb_2023}, fake news dataset by \citet{amjad-etal-2020-data}, Topic Classification dataset, and Universal Dependencies \citep{nivre-etal-2020-universal} are available under CC BY 4.0. 

UQuAD dataset is under CC0-1.0 while UQA \citep{arif-etal-2024-uqa-corpus} is under CC BY 4.0. XSUMUrdu \citep{munaf2023low} summarization dataset is also under CC BY 4.0 license. Paraphrasing dataset is under MIT license. Dakshina dataset \citep{roark-etal-2020-processing} for transliteration is under CC BY-SA 4.0 and OPUS-100 \citep{zhang-etal-2020-improving} \citep{tiedemann-2012-parallel} for translation is under GPL-3.0 license. UrduAssistant \citep{urduassistant} dataset has a MIT license.

All datasets used in this paper comply with their respective license.

\section{Model Size and Budget}
We fine-tuned \texttt{XLM-Roberta-large} for classification tasks which has 550 million parameters. We fine-tuned \texttt{mT5-large} for generation tasks which has 1.2 billion parameters. \texttt{Llama-3-8b} has 8 billion parameters and is fine-tuned for both generation and classification tasks.

Nvidia A100 80GB, Nvidia A100 40GB and Nvidia RTX 6000Ada 48GB were used for fine-tuning of the models. Infernece was done on Nvidia RTX 6000Ada 48GB and Nvidia RTX 4090 24GB. Total GPU time was approximately 200 hours.

\section{Human Annotators}
There are two human annotators in this study: one is the author of this paper (Computer Science graduate), and the other is a research intern (Computer Science senior).
Both annotators are native speakers of Urdu from Pakistan. The research intern was informed about how the data would be used for the evaluation of LLMs for Urdu.

\section{Dataset Size}
\label{sec:appendix-dataset}
This section provides details about the datasets used for the classification and generation tasks evaluated in this study. The test dataset sizes for each task are summarized in the tables below.
\subsection{Classification}
Table \ref{fig:dataset-size} shows the test dataset sizes for classification tasks including Sentiment Analysis, Abuse Detection, Sarcasm Detection, Fake News Detection, Topic Classification, Part-of-Speech (PoS) Tagging, and Named Entity Recognition (NER) Tagging.
\begin{table}[!ht]
    \begin{center}    
    \begin{tabularx}{\columnwidth}{|X|X|}
        \hline
        \textbf{Task} & \textbf{Test Size} \\
        \hline
        Sentiment Analysis & 1000 \\
        \hline

        Abuse Detection & 567 \\
        \hline

       Sarcasm Detection & 1000 \\
        \hline

        Fake News Detection & 130 \\
        \hline

        Topic Classification & 1000 \\
        \hline

        PoS Tagging & 1000 \\
        \hline

        NER Tagging & 1000 \\
        \hline
    \end{tabularx}
    \caption{Test dataset size for each classification task}
    \label{fig:dataset-size}
    \end{center}
\end{table}

\subsection{Generation}
Table \ref{fig:dataset-size-1} shows the test dataset sizes for generation tasks including Summarization, Paraphrasing, Transliteration, English to Urdu Translation, and Urdu to English Translation.
\begin{table}[!ht]
    \begin{center}    
    \begin{tabularx}{\columnwidth}{|X|X|}
        \hline
        \textbf{Task} & \textbf{Test Size} \\
        \hline
        Question Answering & 1000 \\
        \hline

        Summarization & 568 \\
        \hline

       Paraphrasing & 1000 \\
        \hline

        Transliteration & 1000 \\
        \hline

        Translation (en-ur) & 1000 \\
        \hline

        Translation (ur-en) & 1000 \\
        \hline

        AI Assistant & 50 \\
        \hline
    \end{tabularx}
    \caption{Test dataset size for each generation task}
    \label{fig:dataset-size-1}
    \end{center}
\end{table}

\section{Reproducibility and Hyperparameter}
\label{sec:appendix-hyperparam}
The table \ref{fig:lora-config} presents the Lora configuration that we use for fine-tuning \texttt{Llama}.

\begin{table}[!ht]
    \small
    \begin{center}    
    \begin{tabularx}{\columnwidth}{|X|X|}
        \hline
        \textbf{Rank} & 16 \\
        \hline

        \textbf{Alpha} & 16 \\
        \hline

        \textbf{Target Modules} & \texttt{q\_proj, k\_proj, v\_proj, o\_proj, down\_proj, up\_proj, gate\_proj} \\
        \hline
    \end{tabularx}
    \caption{Lora config for \texttt{Llama} fine-tuning}
    \label{fig:lora-config}
    \end{center}
\end{table}

The table \ref{fig:hyper-params} presents the training parameters, including the number of epochs and batch sizes, we use for fine-tuning \texttt{XLM-R}, \texttt{mT5}, and \texttt{Llama}.

\begin{center}
\begin{table}[htbp]
\small
\centering
\caption{Training Parameters for Different Models. E represents number of epochs and B represents the batch size.}
\renewcommand{\arraystretch}{1}
\begin{tabular}{lcccccc}
\toprule
& \multicolumn{2}{c}{\texttt{XLM-R}} & \multicolumn{2}{c}{\texttt{mT5}} & \multicolumn{2}{c}{\texttt{Llama}} \\
\cmidrule(lr){2-3} \cmidrule(lr){4-5} \cmidrule(lr){6-7}
Task & E & B & E & B & E & B \\
\midrule
Sentiment Analysis & 1 & 16 & - & - & 4 & 1 \\
Abuse Detection & 4 & 16 & - & - & 3 & 8 \\
Sarcasm Detection & 4 & 16 & - & - & 3 & 8 \\
Fake News Detection & 6 & 8 & - & - & 3 & 1 \\
Topic Classification & 4 & 32 & - & - & 3 & 8 \\
POS Tagging & 7 & 16 & - & - & 1 & 4 \\
NER Tagging & 10 & 16 & - & - & 3 & 4 \\
Question Answering & - & - & 4 & 8 & 1 & 1 \\
Summarization & - & - & 1 & 4 & 2 & 1 \\
Paraphrasing & - & - & 3 & 3 & 4 & 4 \\
Transliteration & - & - & 5 & 8 & 2 & 4 \\
Translation-en-ur & - & - & 2 & 8 & 1 & 8 \\
Translation  & - & - & 3 & 8 & 1 & 8 \\
AI Assistant  & - & - & 2 & 8 & 2 & 8 \\
\bottomrule
\end{tabular}
\label{tab:training_parameters}
\label{fig:hyper-params}
\end{table}
\end{center}

\section{Human and LLM Evaluation}
\label{sec:appendix-human-eval}
\begin{figure*}[t]
\begin{center}
\centerline{\includegraphics[width=\textwidth]{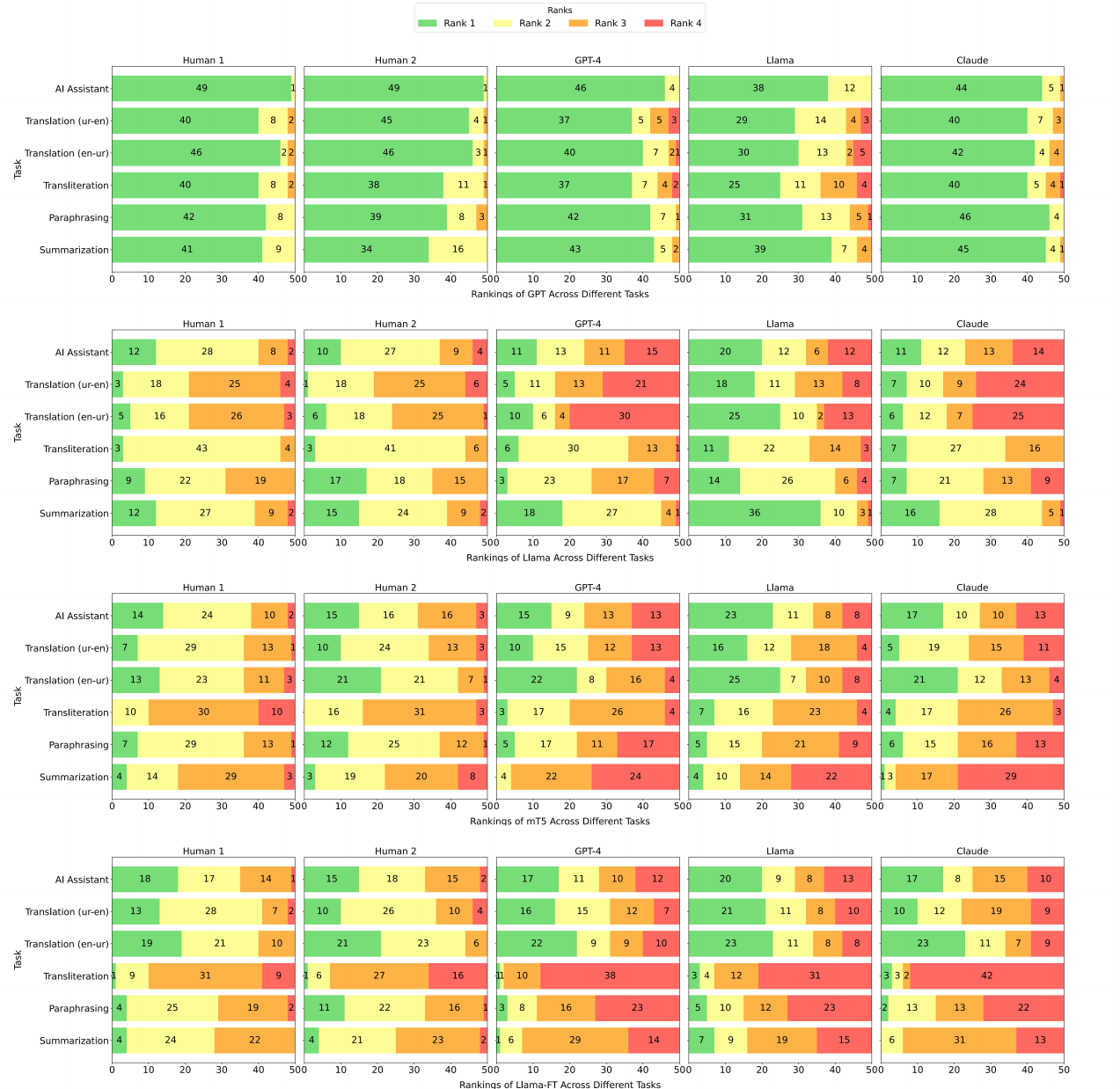}}
\caption{Rank counts of models for all generation tasks.}
\label{fig:all_ratings}
\end{center}
\end{figure*}

Figure \ref{fig:all_ratings} presents the rank distribution for all models across all generation tasks, as evaluated by two human evaluators: \texttt{GPT-4-Turbo}, \texttt{Llama-3-8b}, and \texttt{Claude 3.5 Sonnet}.

\section{Prompts and Evaluation Criteria}
\label{sec:appendix-prompt-criteria}
Table \ref{fig:classprompts} contains the CoT prompts used for the classification tasks. The prompts for classification without CoT are the same, except that the reasoning is not generated, and only the label is required in the output. Table \ref{fig:genprompts} contains the prompts used for the generation tasks. Table \ref{fig:eval-criteria} presents the criteria given to human evaluators, \texttt{GPT} and \texttt{Llama} for assessing the quality of outputs for summarization, paraphrasing, transliteration, English to Urdu translation, and Urdu to English translation. For each evaluation, the LLMs are asked to reason the answer to improve the scoring of the outputs.

\clearpage
\onecolumn

\begin{center}
\begin{longtable}{| p{.20\textwidth} | p{.75\textwidth} |}
\hline
\textbf{Classification Task} & \textbf{CoT Prompt} \\
\hline
\endhead

Sentiment Analysis & \texttt{You are an Urdu sentiment classifier. The input text should be labeled according to its sentiment. The label list is:
['positive', 'negative'] \newline
Use Chain-of-Thought (cot) to reason your answer. \newline ALWAYS RETURN JSON OBJECT IN FOLLOWING FORMAT ONLY: \newline
\{"cot": ..., "label": ...\}}\\
\hline
Abuse Detection & \texttt{You are an Urdu abuse detector. The input text should be labeled according to whether it is abusive or not. The label list is:
['abusive', 'not abusive'] \newline
Use Chain-of-Thought (cot) to reason your answer. \newline
ALWAYS RETURN JSON OBJECT IN FOLLOWING FORMAT ONLY: \newline
\{"cot": ..., "label": ...\}} \\
\hline
Sarcasm Detection & \texttt{You are an Urdu sarcasm detector. The input text should be labeled according to whether it is sarcastic or not. The label list is:
['sarcastic', 'not sarcastic'] \newline
Use Chain-of-Thought (cot) to reason your answer. \newline ALWAYS RETURN JSON OBJECT IN FOLLOWING FORMAT ONLY: \newline
\{"cot": ..., "label": ...\}} \\
\hline
Fake News Detection & \texttt{You are an Urdu fake news detector. The input text should be labeled according to whether it is fake news or not. The label list is:
['fake news', 'not fake news'] \newline
Use Chain-of-Thought (cot) to reason your answer.  \newline ALWAYS RETURN JSON OBJECT IN FOLLOWING FORMAT ONLY: \newline
\{"cot": ..., "label": ...\}} \\
\hline
Topic Classification & \texttt{You are an Urdu topic classifier. The input text should be assign a label from the label list:
['business', 'entertainment', 'health', 'politics', 'science', 'sports', 'world', 'other'] \newline
Use Chain-of-Thought (cot) to reason your answer. \newline ALWAYS RETURN JSON OBJECT IN FOLLOWING FORMAT ONLY: \newline
\{"cot": ..., "label": ...\}} \\
\hline
PoS Tagging & \texttt{You are an Urdu part-of-speech tagger. The word wrapped in <hl> tag should be assigned a PoS tag from the label list:
['noun', 'punctuation mark', 'adposition', 'number', 'symbol', 'subordinating conjunction', 'adjective', 'particle', 'determiner', 'coordinating conjunction', 'proper noun', 'pronoun', 'other', 'adverb', 'interjection', 'verb', 'auxiliary verb'] \newline
Use Chain-of-Thought (cot) to reason your answer. \newline ALWAYS RETURN JSON OBJECT IN FOLLOWING FORMAT ONLY: \newline
\{"cot": ..., "label": ...\}} \\
\hline
NER Tagging & \texttt{You are an Urdu named entity recognizer. The word wrapped in <hl> tag should be assigned a NER tag from the label list:
['time', 'person', 'organization', 'number', 'location', 'designation', 'date', 'other'] \newline
Use Chain-of-Thought (cot) to reason your answer.  \newline ALWAYS RETURN JSON OBJECT IN FOLLOWING FORMAT ONLY: \newline
\{"cot": ..., "label": ...\}} \\
\hline
\caption{The table presents the CoT prompts used for classification tasks.}
\label{fig:classprompts}
\end{longtable}
\end{center}

\begin{center}
\begin{longtable}{| p{0.20\textwidth} | p{0.75\textwidth} |} % Adjust column widths if necessary
\hline
\textbf{Generation Task} & \textbf{Prompt} \\
\hline
\endhead % Repeat header at top of new page

Summarization & \texttt{You are a text summarizer. Your task is to summarize the given Pakistani Urdu text in 1 to 2 sentences. The summary should be in Urdu. \newline ALWAYS RETURN JSON OBJECT IN FOLLOWING FORMAT ONLY: \newline \{"summary": ...\}} \\
\hline
Paraphrasing & \texttt{You are a text paraphraser. Your task is to paraphrase the given Pakistani Urdu text. The paraphrased text should be in Urdu. \newline ALWAYS RETURN JSON OBJECT IN FOLLOWING FORMAT ONLY: \newline \{"paraphrase": ...\}} \\
\hline
Transliteration & \texttt{You are a machine transliterator. Your task is to transliterate the given Pakistani Urdu text to Roman Urdu. \newline ALWAYS RETURN JSON OBJECT IN FOLLOWING FORMAT ONLY: \newline \{"transliteration": ...\}} \\
\hline
Translation (en-ur) & \texttt{You are a machine translator. Your task is to translate the given English text to Pakistani Urdu. \newline ALWAYS RETURN JSON OBJECT IN FOLLOWING FORMAT ONLY: \newline \{"translation": ...\}} \\
\hline
Translation (ur-en) & \texttt{You are a machine translator. Your task is to translate the given Pakistani Urdu text to English. \newline ALWAYS RETURN JSON OBJECT IN FOLLOWING FORMAT ONLY: \newline \{"translation": ...\}} \\
\hline
\caption{The table presents the prompts used for generation tasks.}
\label{fig:genprompts}
\end{longtable}
\end{center}

\begin{center}
\begin{longtable}{| p{.20\textwidth} | p{.75\textwidth} |}
\hline
\textbf{Task} & \textbf{Prompt/Criteria} \\
\hline
\endhead % Repeat header at top of new page
Summarization & \texttt{You are a Pakistani Urdu language expert tasked with evaluating the quality of the summary produced by the summarization model. Score the given output with respect to the given input on a continuous score from 0 to 100 based on the following criteria: \newline
    1. Includes main points and key information from the original text \newline
    2. No grammatical errors \newline
    3. Conveys information in a brief manner \newline
    4. Gives correct information based on the original text \newline
    Think step by step and use reasoning.
    ALWAYS RETURN JSON OBJECT IN THE FOLLOWING FORMAT ONLY: \newline
    \{"reasoning": ..., "score": ...\}}\\
\hline
Paraphrasing & \texttt{You are a Pakistani Urdu language expert tasked with evaluating the quality of paraphrased text produced by the paraphrasing model. Score the given output with respect to the given input on a continuous score from 0 to 100 based on the following criteria: \newline
    1. Retains the original meaning and key ideas \newline
    2. No grammatical errors \newline
    3. Use of different words and phrases than the original text \newline
    Think step by step and use reasoning. ALWAYS RETURN JSON OBJECT IN THE FOLLOWING FORMAT ONLY: \newline
    \{"reasoning": ..., "score": ...\}} \\
\hline
Transliteration & \texttt{You are a Pakistani Urdu language expert tasked with evaluating the quality of transliterated text produced by the transliteration model. Score the given output with respect to the given input on a continuous score from 0 to 100 based on the following criteria: \newline
    1. Correctness of words (keeping in mind that different words can have different spelling) \newline
    2. Proper capitalization of words \newline
    Think step by step and use reasoning. ALWAYS RETURN JSON OBJECT IN THE FOLLOWING FORMAT ONLY: \newline
    \{"reasoning": ..., "score": ...\}} \\
\hline
Translation (en-ur) & \texttt{You are a Pakistani Urdu language expert tasked with evaluating the quality of translated text produced by the translation model. Score the given output with respect to the given input on a continuous score from 0 to 100 based on the following criteria: \newline
    1. Conveys the meaning of the original text without omissions or additions \newline
    2. No grammatical errors \newline
    3. Retains the style and tone of the original text \newline
    Think step by step and use reasoning. ALWAYS RETURN JSON OBJECT IN THE FOLLOWING FORMAT ONLY: \newline
    \{"reasoning": ..., "score": ...\}} \\
\hline
Translation (ur-en) & \texttt{You are a Pakistani Urdu language expert tasked with evaluating the quality of translated text produced by the translation model. Score the given output with respect to the given input on a continuous score from 0 to 100 based on the following criteria: \newline
    1. Conveys the meaning of the original text without omissions or additions \newline
    2. No grammatical errors \newline
    3. Retains the style and tone of the original text \newline
    Think step by step and use reasoning. ALWAYS RETURN JSON OBJECT IN THE FOLLOWING FORMAT ONLY: \newline
    \{"reasoning": ..., "score": ...\}} \\
\hline
AI Assistant & \texttt{Please act as an impartial judge and evaluate the quality of the response provided by an AI assistant to the user question displayed below. Your evaluation should consider factors such as the helpfulness, relevance, accuracy, depth, creativity, and level of detail of the response. Begin your evaluation by providing a short explanation. Be as objective as possible. After providing your explanation, please rate the response on a scale of 0 to 100. ALWAYS RETURN JSON OBJECT IN THE FOLLOWING FORMAT ONLY:\newline
\{"reasoning": ..., "score": ...\}} \\
\hline
\caption{The table presents the criteria used to evaluate the outputs of the generation task in the form of an LLM prompt.}
\label{fig:eval-criteria}
\end{longtable}
\end{center}

\end{document}